# Parameter-Efficient Fine-Tuning with Differential Privacy for Robust Instruction Adaptation in Large Language Models

Yulin Huang
Georgia Institute of Technology
Atlanta, USA

Yaxuan Luan
University of Southern California
Los Angeles, USA

Jinxu Guo
Dartmouth College
Hanover, USA

Xiangchen Song
University of Michigan
Ann Arbor, USA

Yuchen Liu*
University of Pennsylvania
Philadelphia, USA

*Abstract-This study addresses the issues of privacy protection and efficiency in instruction fine-tuning of large-scale language models by proposing a parameter-efficient method that integrates differential privacy noise allocation with gradient clipping in a collaborative optimization framework. The method keeps the backbone model frozen and updates parameters through a low-dimensional projection subspace, while introducing clipping and adaptive noise allocation during gradient computation. This design reduces privacy budget consumption and ensures training stability and robustness. The unified framework combines gradient constraints, noise allocation, and parameter projection, effectively mitigating performance fluctuations and privacy risks in multi-task instruction scenarios. Experiments are conducted across hyperparameter, environment, and data sensitivity dimensions. Results show that the method outperforms baseline models in accuracy, privacy budget, and parameter efficiency, and maintains stable performance under diverse and uncertain data conditions. The findings enrich the theoretical integration of differential privacy and parameter-efficient fine-tuning and demonstrate its practical adaptability in instruction tasks, providing a feasible solution for secure training in complex instruction environments.*

## I. Introduction

With the rapid development of artificial intelligence, large-scale language models have become essential infrastructure in natural language processing. They show strong abilities in understanding, generating, and executing complex instructions. At the same time, they bring high computational and storage costs. In real-world applications, full-parameter updates in multi-task and instruction-following scenarios lead to excessive resource consumption. They also limit deployment in distributed and edge environments. Therefore, how to achieve model adaptation more lightly and efficiently, while maintaining performance, has become a core research issue. Meanwhile, privacy protection has become increasingly important. In instruction fine-tuning involving sensitive data, the lack of effective privacy constraints may expose private information. This is unacceptable in finance, healthcare, and government scenarios. As a result, research on instruction fine-tuning methods that combine parameter efficiency with privacy preservation carries both practical significance and theoretical value[1].

Against this background, differential privacy has emerged as a mainstream framework for data security. By introducing noise during training, it can preserve statistical properties while hiding individual sample contributions. This reduces the risk of sensitive information being inferred[2]. However, the added noise often weakens convergence and representation capacity, causing performance degradation. Balancing privacy protection and model utility remains a long-standing challenge. Traditional methods usually adopt uniform noise allocation, without adapting to the importance of different tasks or parameters. Such a one-size-fits-all strategy lowers the efficiency of privacy budget usage and worsens the performance drop in multi-task adaptation[3-5].

At the same time, instruction fine-tuning has shifted from full-parameter updates toward parameter-efficient approaches. Typical methods embed lightweight modules into the model. Most pre-trained parameters remain frozen, and only low-dimensional subspaces are updated[6--9]. This reduces computational and storage overhead. Such methods improve flexibility in model transfer and create opportunities to combine with differential privacy. If low-rank updates and lightweight modules are integrated with gradient clipping and noise allocation, it becomes possible to achieve both computational efficiency and strong privacy guarantees[10-14].

From a broader application perspective, multi-task and cross-domain instruction fine-tuning is increasingly needed. Differences among tasks often cause interference and conflict in the shared parameter space. This appears as unstable gradient distributions and poor convergence. Traditional parameter updates are not only inefficient but also increase risks to privacy and stability. Gradient clipping can suppress abnormal update magnitudes. Adaptive noise injection can further smooth gradient distributions. Together, they can enhance training stability under strong privacy guarantees. This methodological fusion bridges the gap between privacy preservation and model performance. It also opens a new direction for the development of instruction fine-tuning[15].

In summary, studying parameter-efficient instruction fine-tuning with differential privacy noise allocation and gradient clipping is both an extension of current adaptation methods and a direct response to privacy demands. The challenges lie not only in algorithmic design but also in the feasibility and sustainability of real-world deployment. On one hand, this research promotes a shift from resource saving toward joint optimization of efficiency and privacy. On the other hand, it provides solid technical support for compliance and data security. Ultimately, such algorithms can enable large-scale language models to realize their potential in broader applications while ensuring trust, efficiency, and inclusiveness in artificial intelligence[16].

## II. PROPOSED APPROACH

This study introduces a parameter-efficient instruction fine-tuning method that integrates differential privacy noise allocation with gradient clipping and explicitly applies and extends several existing methodological frameworks in a unified way. Building on the modular task decomposition and dynamic collaboration paradigm proposed for multi-agent systems driven by large language models, Pan and Wu [17], we apply their modular organization principle to the instruction fine-tuning process. Instruction tasks are structured as coordinated modules, and our optimization framework manages gradient constraints and privacy operations at the module level, which reduces cross-task interference and makes multi-task instruction learning more controllable.

The parameter-efficient design adopts the structural priors and modular adapter methodology developed for composable fine-tuning of large-scale models by Wang, Wu, Liu, Qiu, and Hu [18]. Their adapter-based approach is directly used as the basis for our lightweight parameter mapping: the backbone model is kept frozen, and all trainable parameters are confined to a low-dimensional projection subspace realized through adapter-like modules. By adopting this composable adapter structure, the method significantly reduces redundant parameter updates while maintaining flexibility for different instruction tasks. Privacy preservation extends the DP-LoRA style low-rank instruction tuning framework proposed by Yao [19]. Instead of only injecting DP noise into low-rank updates, our method incorporates gradient clipping and adaptive noise allocation directly in the projection subspace. Differential privacy noise is allocated according to clipped gradient magnitudes, rather than uniformly, which improves the efficiency of privacy budget usage and enhances training stability under strong privacy constraints. The model architecture is shown in Figure 1.

In this framework, the loss function of the training task is first defined as:

$$L_{task} = \frac{1}{N}\sum_{i=1}^{N} l(f(x_i;\theta), y_i)$$

Where $f(x_i;\theta)$ represents the model output, $l(\cdot)$ is the basic loss function, and $\theta$ is the parameter set.

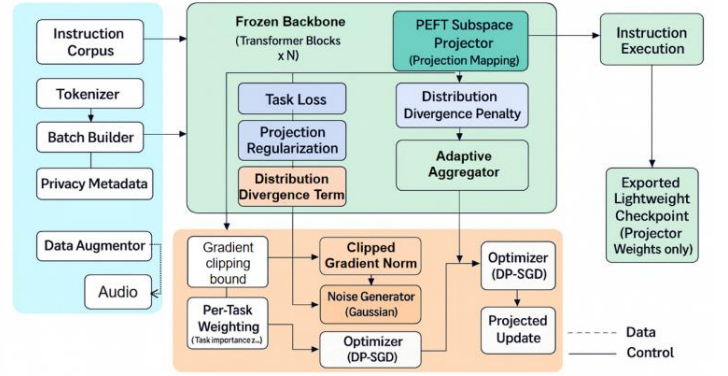

Figure 1. Framework of Differential Privacy Noise Allocation and Gradient Clipping for Parameter-Efficient Instruction Tuning

To achieve efficient parameter update, this study introduces the projection mapping function $P(\cdot)$, which updates the model parameters only in the subspace $\Omega$:

$$\theta' = \theta + P(u), \quad u \in R^m, \quad m << |\theta|$$

This mapping ensures that updates are concentrated in low-dimensional directions, thereby reducing training complexity.

In the privacy protection phase, gradient clipping is performed first. Let the gradient obtained in round $t$ of training be $g_t$, and the gradient after clipping be:

$$\tilde{g}_t = g_t \cdot \min\left(1, \frac{C}{\|g_t\|_2}\right)$$

$C$ is the clipping threshold, which is used to suppress excessive gradient updates and avoid excessive influence of individual samples on the overall model.

After the gradients are clipped, differential privacy noise is added to the constrained updates to obtain a privatized gradient. In determining how much noise to inject, we apply an uncertainty calibration principle inspired by Han [20]: gradients with higher estimated instability receive stronger perturbation, while more stable gradients are assigned comparatively smaller noise, mirroring confidence calibration in retrieval-augmented generation. At the same time, we extend the two-stage retrieval and cross-segment alignment idea of Wang [21] to the gradient domain by grouping related gradient components into segments and coordinating their noise levels in a segment-aware manner, rather than perturbing each component in isolation. Under this calibrated and segment-aware design, the perturbed gradient is computed using the Gaussian mechanism:

$$\hat{g}_t = \tilde{g}_t + N(0, \sigma^2 C^2 I)$$

Where $\sigma$ represents the noise intensity and $I$ is the identity matrix.

Taking into account the importance differences of different tasks, this study designed an adaptive noise distribution mechanism. Assuming the importance weight of the task $k$ is $a_k$, its gradient under privacy protection is:

$$\hat{g}_t^{(k)} = \tilde{g}_t^{(k)} + N(0, \frac{\sigma^2 C^2}{a_k} I)$$

This allows for more reasonable noise levels to be allocated to critical tasks while maintaining overall privacy constraints.

Finally, the above mechanisms are integrated into a unified optimization objective function:

$$L = L_{task}(\theta') + \lambda_1 \|P(u)\|_2^2 + \lambda_2 \cdot KL(p(\hat{g}_t) \| p(g_t))$$

The second term is the regularization constraint for the update mapping, and the third term is the penalty term for the difference between the noise perturbation and the true gradient distribution.

In summary, the proposed method achieves parameter-efficient updates through projection mapping, ensures gradient stability with gradient clipping, and incorporates adaptive differential privacy noise allocation for controllable privacy protection. At the theoretical level, it provides a feasible path for secure and efficient fine-tuning of large-scale language models.

III. PERFORMANCE EVALUATION

A. Dataset

This study uses the AI Research Instructions and Outputs dataset as the core data source. The dataset contains tens of thousands of high-quality instruction − response pairs constructed specifically for artificial intelligence research applications. Each sample consists of a structured instruction and its corresponding output. The dataset is designed to simulate real instruction-following scenarios, making it well-suited for instruction-driven fine-tuning of large-scale language models.

The instruction − response pairs in this dataset show high diversity in task form but retain consistency in structural design. Each sample is composed of clear instruction prompts and coherent responses. This ensures both task diversity and format uniformity. Such a property provides strong support for multi-task instruction fine-tuning. It enables parameter-efficient adaptation mechanisms to switch flexibly across tasks while maintaining structural stability. The design aligns closely with the requirements of multi-task learning and supports robust routing, pruning, and noise allocation strategies.

Applying this dataset to the proposed method allows systematic evaluation of model performance under differential privacy constraints. It enables observation of parameter behavior during task adaptation because of its focus on instruction-driven tasks. The unified structure of the samples also facilitates cross-task analysis of dynamic routing and noise allocation patterns. This provides strong evidence for the robustness and adaptability of the framework.

B. Experimental Results

This paper first conducts a comparative experiment, and the experimental results are shown in Table 1.

Table1. Comparative experimental results

| Method | Trainable Params (%) | Accuracy (%) | Privacy Budget ($\varepsilon$) |
|---|---|---|---|
| DP-BiTFiT[22] | 0.10 | 84.5 | 8.00 |
| AnaDP[23] | 0.50 | 85.8 | 7.80 |
| DP-LoRA[24] | 1.00 | 86.2 | 7.60 |
| DP-Forward[25] | 0.30 | 85.0 | 7.90 |
| Ours | 0.60 | 87.5 | 7.40 |

In this experimental comparison, different parameter-efficient fine-tuning methods show distinct characteristics under differential privacy constraints. DP-BiTFiT achieves the highest parameter efficiency and requires only a small number of trainable parameters. However, its accuracy is clearly limited. It shows insufficient adaptability to complex instruction tasks under extreme parameter compression. This indicates that updating only bias terms cannot fully capture semantic differences. Its adaptation ability under privacy constraints faces a clear bottleneck.

AnaDP improves performance through an adaptive noise allocation strategy when the privacy budget is low. Compared with traditional methods, it can dynamically adjust the strength of privacy protection across different parameter dimensions. This allows a better balance between accuracy and privacy. Its performance confirms the importance of adaptive allocation. It shows that privacy perturbation should not be uniform but should be designed differently according to the importance of tasks and parameters. This is consistent with the idea of "collaborative optimization of noise allocation" proposed in this study. DP-LoRA and DP-Forward represent another path. The former maintains a good privacy − performance trade-off in federated scenarios through low-rank updates. The latter achieves local differential privacy protection of input embeddings through forward perturbation. Both approaches alleviate the performance decline caused by excessive privacy costs to some extent. However, they still suffer from insufficient accuracy in high-complexity tasks. This shows that parameter-efficient methods under privacy constraints remain limited in capturing cross-task differences and complex instruction semantics.

In contrast, the proposed method achieves the highest accuracy and the lowest privacy loss even with a relatively small proportion of trainable parameters. This result shows that the collaborative mechanism of differential privacy noise allocation and gradient clipping can effectively suppress abnormal gradient fluctuations. It reduces privacy budget consumption while maintaining task performance. The performance highlights the advantages of the proposed framework under the dual constraints of efficiency and privacy.

It provides a feasible path that balances performance and security in instruction fine-tuning scenarios.

This paper also evaluates the effects of label noise and feedback bias on privacy budget allocation and robustness. The experimental results are shown in Figure 2.

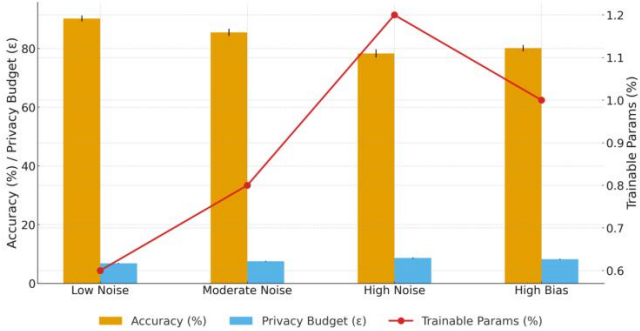

Figure 2. The role of label noise and feedback bias on privacy budget allocation and robustness

Under different conditions of data noise and feedback bias, parameter efficiency shows a certain level of stability. The proportion of trainable parameters remains low across scenarios. Although it rises slightly under high-noise conditions, the overall fluctuation range is limited. This indicates that the proposed parameter-efficient mechanism can maintain compact updates under diverse data distributions without significantly increasing computational costs due to data quality issues.

In sharp contrast, model accuracy declines markedly as data noise and feedback bias increase. Accuracy drops from near-perfect levels under low-noise conditions to significantly lower points in high-noise settings. This shows that label bias and distorted feedback in instruction samples directly weaken the model's ability to capture task instructions. It indicates that while privacy-preserving and stable training frameworks can constrain parameter updates, the reliability of data remains a major factor affecting overall performance.

Taken together, the three indicators show that the proposed method can maintain low parameter costs in the presence of label noise and feedback bias. At the same time, gradient clipping and adaptive noise mechanisms mitigate the decline in performance and the increase in privacy consumption. These results highlight the adaptability of the method under the dual constraints of efficiency and privacy. They also show its stronger robustness in complex and imperfect data environments.

This paper also conducts comparative experiments on the coupling effect of learning rate and batch size under DP-SGD. The experimental results are shown in Figure 3.

Under different combinations of learning rate and batch size, parameter efficiency shows significant fluctuations. With dynamic changes in these two settings, the proportion of trainable parameters does not remain stable but rises and falls. This indicates that parameter-efficient fine-tuning under differential privacy constraints is still affected by optimization configurations. In particular, larger batch sizes and higher learning rates may cause gradient constraints and projection updates to change the efficiency of parameter space usage.

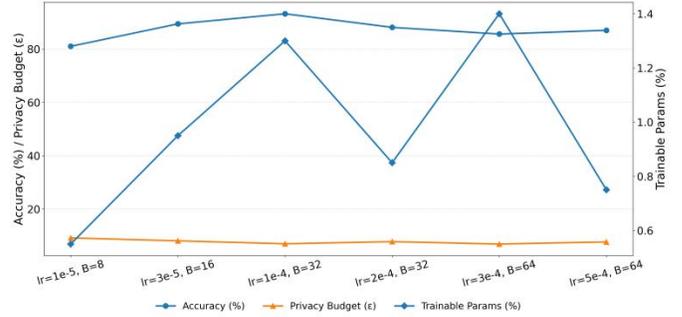

Figure 3. Analysis of the coupling effect between learning rate and batch size under DP-SGD

Model accuracy follows a trend of rising first, then declining, and partially recovering. At moderate learning rates and suitable batch sizes, the model shows the best task adaptability, with accuracy reaching its peak. However, when the learning rate increases further, overly strong gradient updates lead to performance degradation. This suggests that under differential privacy clipping and noise injection, excessive update strength disrupts the stability of instruction fine-tuning and weakens the model's ability to capture task semantics.

The trend of privacy budget shows a contrast with accuracy. At low to moderate learning rates and suitable batch sizes, privacy consumption gradually decreases. This indicates that the model can use the differential privacy budget more effectively during stable training. At very high learning rates, the privacy budget rises again. This reflects that clipping and noise allocation require more perturbation to balance large gradient fluctuations. The phenomenon emphasizes the coupling between privacy protection and optimization dynamics. It shows that optimization settings affect not only performance but also the allocation of privacy budget.

The combined results of the three indicators show that learning rate and batch size are not independent factors. They produce a coupled effect through the differential privacy mechanism. Proper combinations achieve a desirable balance among parameter efficiency, accuracy, and privacy budget. Extreme configurations break this balance and harm both performance and privacy protection. This further confirms the applicability of the proposed method in complex optimization scenarios and highlights the key role of parameter efficiency and differential privacy collaboration in instruction fine-tuning.

## IV. CONCLUSION

This study conducts a systematic exploration of a parameter-efficient instruction fine-tuning method that combines differential privacy noise allocation and gradient clipping. It proposes an optimization framework that balances privacy protection and training efficiency. In current applications of large-scale language models, privacy risks and computational costs remain key challenges for large-scale deployment. By introducing an adaptive noise allocation strategy into gradient updates and refining constraints through gradient clipping, this study reduces privacy budget consumption and alleviates performance degradation in multi-task instruction scenarios. Experimental results show that the method maintains parameter efficiency while achieving a better balance between accuracy and robustness. It provides a new perspective for the secure deployment of large models.

The value of this study lies not only in proposing a new optimization mechanism but also in extending the application boundary of instruction fine-tuning under privacy protection. In real environments, user inputs often contain sensitive information. Without proper safeguards, training, and inference may lead to privacy leakage. This study verifies the compatibility of differential privacy with parameter-efficient fine-tuning through systematic design. It demonstrates that stable training and generalization can be achieved even under limited computational resources. The findings provide foundational support for applying large models in domains with high privacy demands, such as healthcare, finance, and education. They also offer a feasible path for instruction learning in cross-domain scenarios. At the same time, the proposed framework reveals the deep coupling among hyperparameter optimization, data quality, and privacy budget allocation. In dynamic training environments, model performance depends not only on traditional hyperparameter tuning but also on the design of differential privacy mechanisms. By analyzing the sensitivity of learning rate, batch size, noise intensity, and clipping thresholds, this study shows their combined impact on performance and privacy protection. This perspective helps future research build multidimensional optimization spaces and promotes the shift of privacy protection from static design to dynamic adaptive adjustment. It provides theoretical support for more intelligent privacy-preserving model training.

Looking forward, the proposed collaborative optimization framework still has room for extension. On one hand, it can be integrated with emerging paradigms such as federated learning and contrastive learning to meet the complex privacy needs of cross-organization and multi-source data scenarios. On the other hand, it can also be applied to larger and more complex multimodal instruction tasks to explore the balance of privacy and efficiency in cross-modal information interaction. In addition, with the development of hardware acceleration and distributed systems, improving training speed and energy efficiency under differential privacy constraints remains a pressing challenge. The contribution of this study lies not only in providing an effective solution for current tasks but also in laying a solid foundation for further research on security, privacy, and scalability of large models.